# Scalable and consistent few-shot classification of survey responses using text embeddings

Jonas Timmann Mjaaland[a,1], Markus Fleten Kreutzer[a,1], Halvor Tyseng[a,1], Rebeckah K. Fussell[c], Gina Passante[d], N.G. Holmes[c], Anders Malthe-Sørenssen[a,b], and Tor Ole B. Odden[b,2]

**Qualitative analysis of open-ended survey responses is a commonly-used research method in the social sciences, but traditional coding approaches are often time-consuming and prone to inconsistency. Existing solutions from Natural Language Processing such as supervised classifiers, topic modeling techniques, and generative large language models have limited applicability in qualitative analysis, since they demand extensive labeled data, disrupt established qualitative workflows, and/or yield variable results. In this paper, we introduce a text embedding-based classification framework that requires only a handful of examples per category and fits well with standard qualitative workflows. When benchmarked against human analysis of a conceptual physics survey consisting of 2899 open-ended responses, our framework achieves a Cohen's Kappa ranging from 0.74 to 0.83 as compared to expert human coders in an exhaustive coding scheme. We further show how performance of this framework improves with fine-tuning of the text embedding model, and how the method can be used to audit previously-analyzed datasets. These findings demonstrate that text embedding–assisted coding can flexibly scale to thousands of responses without sacrificing interpretability, opening avenues for deductive qualitative analysis at scale.**

Artificial Intelligence | Natural Language Processing | Automated coding | Few-shot classification

Qualitative analysis is a cornerstone of research in the social sciences. Surveys are a common data source for such analysis, especially free-response surveys featuring open-ended questions. These types of data offer unique opportunities to gather deep insights into phenomena that can go beyond what is possible to capture using quantitative scales or methods.

However, these kinds of data can be challenging to analyze at scale. Because they are most often analyzed "by hand", when free-response survey datasets become large they become increasingly intractable for individual qualitative researchers to analyze, and require increasing amounts of time and resources from research teams (1). Furthermore, researchers may code or analyze such data inconsistently, with judgments varying significantly between different researchers or even the same researcher at different times. These issues with scale and consistency often limit the scope of qualitative studies.

Several techniques from Natural Language Processing (NLP) have been proposed to mitigate these challenges. However, these methods introduce new difficulties, especially when integrated into existing qualitative workflows. For example, supervised machine learning models have been trained to classify unseen texts based on labeled examples (2). These methods fit well with the deductive qualitative traditions in which one applies a set of codes or categories defined a priori to a set of data. However, training a well-performing classifier requires hundreds or even thousands of labeled examples in order to create the necessary training data, making such projects nearly as resource-intensive as a standard qualitative analysis (3). Furthermore, such models can have varying levels of performance on data that differ significantly from their training data, potentially limiting their scope (4).

Methods for topic modeling allow the extraction of latent themes or trends from unstructured textual datasets (5). Many of these methods are unsupervised, meaning that they do not require a priori human analysis to train. Consequently, they fit well into the category of inductive qualitative analysis, in which a researcher analyzes a set of data in order to identify themes or patterns that were not known beforehand. However, these methods also introduce new challenges. For instance, topic models often require researchers to specify the number of topics ahead of time, and it can be difficult to know how many topics make sense in a specific set of data (6). The data needed for topic modeling methods are often large, on the order of hundreds or thousands of texts in order to find robust themes, and such data often

## Significance Statement

In the social sciences, free-response surveys are commonly used to gather deep, qualitative insights and are typically analyzed by hand. However, these approaches do not scale to large datasets and are susceptible to inconsistencies. Here, we introduce a framework for using text embeddings to deductively classify open-ended survey datasets at scale. The framework requires only a handful of pre-labeled examples per category and fits into standard qualitative workflows. It can achieve near-perfect agreement with expert human coders and can be used to audit previously analyzed datasets for inconsistencies. These findings demonstrate that text embedding-assisted coding can be a powerful tool for qualitative analysis, allowing researchers to flexibly and iteratively analyze datasets at previously-intractable scales.

Author affiliations: [a]Center for Interdisciplinary Education, University of Oslo, NO-0316 Oslo, Norway; [b]Center for Computing in Science Education, University of Oslo, NO-0316 Oslo, Norway; [c]Laboratory of Atomic and Solid State Physics, Cornell University, Ithaca, New York 14853, USA; [d]Department of Physics, California State University Fullerton, Fullerton, California 92831, USA require significant filtering and pre-processing to be useable. Such methods often have issues with random initialization, where a small change to an initialization state will produce vastly different topics (7). Finally, the topics found by an article may or may not be of interest to a human analyst, since such techniques primarily look for patterns in word usage which do not always correlate with topics of theoretical interest.

Some researchers have begun to use chat-based LLMs, such as ChatGPT, as a tool for qualitative analysis (8). Here too, however, researchers run into new challenges. Chat-based LLMs are often "black boxes", meaning that researchers have little control over or insight into their behavior (9). The results of such an analysis can also be stochastic, varying from run to run (10), and commercial LLMs are inappropriate for many kinds of human-subjects data due to privacy restrictions (11). Finally, chat-based LLMs use significantly more energy than most other analysis tools (12).

In this paper, we propose a framework for using current NLP tools to perform deductive qualitative survey analysis that addresses these challenges of scalability and consistency, while also fitting with standard qualitative workflows. This framework leverages embeddings, vectorized representations of textual meaning, and can be run based on a handful of carefully-chosen representative responses (few-shot classification). We illustrate this framework by analyzing a dataset of approximately 2900 free-text responses to a survey about the sources of variability in a set of fictional experimental physics scenarios, benchmarking the results against prior human analysis. We show how the framework produces high agreement with human coding when we simulate an exhaustive coding task (in which every response receives a specific code and none are left out of the analysis). Moreover, this performance improves with fine-tuning of the embeddings model. Furthermore, we show how embeddings can be used to audit previously-coded datasets, identifying inconsistencies and edge-cases in the coded data. In this way, this framework combines the intuitive and human-centered aspects of standard deductive qualitative analysis with the speed and scalability of NLP techniques to provide a tool for qualitative researchers to analyze large textual datasets.

## Integrating Qualitative Analysis and Text Embeddings

**The Qualitative Workflow.** The framework we describe here fits into the tradition of deductive qualitative content analysis. That is to say, a researcher uses a framework or set of categories defined a priori to deductively classify each survey response based on the content of that response.

This workflow often proceeds as follows: first, researchers gather, download, and prepare a dataset for review, for example in an excel file or a qualitative analysis program like NVivo, Atlas.ti, or MaxQDA. They next review the data and construct categories based on existing theoretical frameworks, research questions, and initial analysis of the dataset. Researchers will additionally gather a set of illustrative examples that may be used as references for the categories.

The coding scheme is then applied to the dataset, with the qualitative researcher reviewing each response and determining the category into which it best fits. In some cases, coding will be done selectively, where certain responses do not receive a code if they do not fit well with any of the categories and/or are assigned a generic category like *Other*. In other cases coding is done exhaustively, where every response must be coded into one of the primary categories. This process often takes a substantial amount of time, especially with large datasets, and multiple coders may independently code or review a subset of the data to evaluate their level of agreement on the analysis (1). Finally, researchers examine aggregate trends or emergent themes based on the categories assigned.

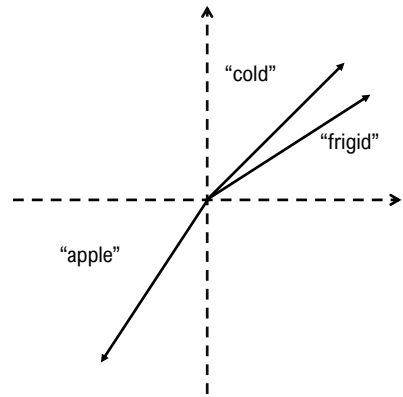

**Fig. 1.** Illustration of how words with similar meaning are closer to each other in the embedding space.

**Text Embeddings.** Text embeddings are vector representations that map the semantic meaning of a text onto a continuous and high-dimensional vector space, where semantically similar pieces of text are located near one another (13) as defined by a distance metric defined for the space. For example, in a well-constructed embedding model, the vectors for the words "cold" and "frigid" would be close to each other in the embedding space and relatively far away from the vector for an unrelated word such as "apple" (Figure 1).

Transformer-based text embeddings are in addition contextual, meaning that the words pay attention to each other. For example, the embedding of the word "bank" will vary depending on whether it is used in the context edixof "river" or "investment" (14). These embeddings are then pooled across text segments to create one context-aware text embedding.

Text embeddings are produced by text embedding models (13). These models learn semantic relationships via a complex training regime, which usually involves self-supervised training on large amounts of text data, with possible supervised fine-tuning for downstream tasks (15). There are a plethora of different text embedding models that are publicly accessible and vary in parameter size, token limit, performance at specific tasks, and training data (16).

**The f-SCUTE, Few-Shot Survey Classification Using Text Embeddings.** Our proposed framework leverages these properties of embeddings in order to perform analysis of open-ended survey responses at scale. The framework has five main steps, which fit with the previously-described qualitative analysis workflow: (1) Data Preparation, (2) Embedding, (3) Category mapping, (4) Classification, (5) Evaluation.

*(1) **Data preparation.*** When using text embedding models, it is important to be as close to the natural text as possible. Unlike many other NLP techniques, text embeddings work best when leaving punctuation, casing, conjunctions, articles, and other linguistic features intact. The survey responses should be in a table with separate rows for each entry and relevant metadata as columns. Data preparation and structure is, therefore, similar to that used in standard qualitative analysis.

*(2) **Embedding.*** The next step is to embed the responses, i.e. convert each response into a numeric vector. To do this, we use open-source pre-trained transformer-based text embedding models. The Hugging Face ecosystem provides access to many such models (17). Models are ranked according to the MTEB (Massive Text Embedding Benchmark), a benchmark that provides a balanced indication of a model's performance on several standard tasks such as classification, clustering, and retrieval (16).

Once the responses are passed through a model, each response is converted into a single *embedding vector* that denotes the location of the response when it is "embedded" in the high-dimensional vector space. The embedded vectors cannot, in general, be compared across different embedding models.

*(3) **Category mapping.*** As in traditional deductive qualitative analysis, a central part of our framework is to manually define a robust coding scheme. This part of the framework proceeds in the same way as in traditional qualitative analysis: a qualitative researcher either brings in an existing coding scheme or defines a new coding scheme for the analysis "by hand," informed by theory, prior research, and in conversation with the data. The coding scheme reflects what researchers are looking for in the data and functions as the conceptual foundation for the analysis.

Once a coding scheme is defined, the next step is to map this coding scheme onto the embedding space. That is, we need to represent a semantic category (code) as a vector within the high-dimensional semantic space defined by the embedding model. We determine the category's embedding vector, by creating a category centroid. This centroid is the average of the embedding vectors of a number of representative example responses for each category chosen by the qualitative researcher based on their alignment with the intended meaning of the category. The centroid acts as its category's proxy in the embedding space. This way of representing a category with examples is similar to that of prototypical networks, which is most commonly used for few-shot classification of images (18).

Traditional qualitative analysis relies on clearly-defined theoretical categories. In the same way, in the present framework, the quality of responses that have been selected to represent categories will directly impact the quality of analysis results. These responses should therefore be carefully chosen by a human researcher, drawing on their interpretive judgment and domain expertise.

*(4) **Classification.*** Once category centroids have been defined, we classify each response by measuring the similarity (or distance) between its embedding vector and each category centroid using a similarity metric. Figure 2 shows how each category centroid is defined by averaging the embedding vectors of several example responses. In mutually exclusive classification, a given response is considered to belong to the closest category measured using the similarity metric between the embedding vector for the response and the category centroids. For multi-label classification, we transform the similarity metric values into a set of confidence scores, using e.g. a softmax function (19). There are several possible ways to define the similarity metric. In this framework, we use cosine similarity, which provides a robust measure of similarities in high-dimensional vector spaces (13, 20).

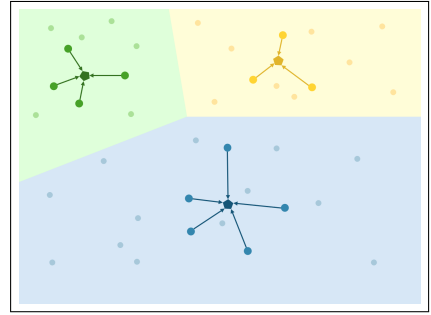

**Fig. 2.** Illustrative figure. Representative responses for each category (bold circles) are averaged to create a category centroid (bold pentagon), which is similar to protoypical networks (18). Responses are categorized based on their proximity to the nearest centroid.

*(5) **Evaluation.*** The final step in the framework is to evaluate results, for example by benchmarking them against human coding of a subset of the dataset, and to examine trends found in the classified data. We note that unlike traditional qualitative analysis, it is simple and fast to adjust categories and reclassify data when using NLP methods in general and the DeFSSCUTE framework in particular. That is, adjustments of the category centroids — such as removing, swapping, adding, or refining representative responses — can be performed quickly and iteratively if initial runs reveal ambiguities.

## Application to a Physics Education Survey

As a proof-of-concept, we test the framework on a large survey dataset from the field of physics education research. Because the dataset has previously been qualitatively analyzed by hand, it provides a useful benchmark for how the proposed framework performs.

**Survey Dataset.** The dataset consists of 2,899 open-ended responses to questions about the sources of variability in a set of fictional experimental physics scenarios (21, 22). Respondents were 753 students enrolled in physics courses at 11 universities in the United States and one in Europe; 441 students were enrolled in introductory physics courses and 312 students were enrolled in upper-division physics courses (such as Quantum Mechanics). The complete demographic and institutional characteristics of the student population are in and Tables S1 and S2 in the SI Appendix.

In the survey, respondents were given information about an experimental scenario, shown a fictional set of collected data, and asked "What is causing the spread in the distribution? List as many causes as you can think of." Students could enter individual sources of variability in separate text boxes. The four scenarios presented were a projectile motion experiment, a Brownian motion experiment, a single-particle single-slit experiment, and a Stern-Gerlach experiment. Every student answered the question about the projectile motion experiment scenario, and upper-division physics students also answered a question about one other experimental scenario.

In the original study, responses were coded by humans based on a coding scheme with four primary categories which were developed emergently through repeated engagement with the data as well as inspiration from the Modeling Framework for Experimental Physics (23). The coding categories are mutually exclusive; that is to say, each response receives a single code. The four codes are defined as follows (see more details in (22)):

- **L = Limitations**: Variability owing to our inability to perfectly model and measure a physical system in a real-world experiment.

- **P = Physical Principles**: Variability inherent in the theoretical abstraction of an experiment or statistical modeling of measurement.

- **S = Statistics**: Reasoning that explains the shape of the distribution through a statistical or data-driven lens, with no explicit reasoning about the physical scenario or apparatus.

- **O = Other**: Non-empty responses that could not be categorized into the three other codes above.

**Table 1. The number of responses coded as belonging to each of the four categories.**

| Category | No. of responses |
|---|---|
| **Limitations** | 2228 |
| **Physics Principles** | 161 |
| **Statistics** | 102 |
| **Other** | 408 |
| **Total** | 2899 |

This coding scheme was initially applied to the data by two independent researchers, each coding a subset of the full dataset. A random sample of 40 responses was coded by both researchers to test for inter-rater reliability and found almost perfect agreement (a Cohen's Kappa of 0.85) (22). A single human coder then categorized the remaining responses. Table 1 shows the distribution of the different codes within the dataset according to the human coders.

**Applying the Framework.** To apply the proposed framework to this survey data, we begin with step one, preparing the data. Text embedding models process natural language, and the raw responses were therefore already in a format suitable for embedding, allowing us to proceed directly to step two, embedding.

To embed the data, we first selected six different open-source text embedding models (24–29). Some of these models allow users to give instructions or even pick from a range of variants to fit a specific task (e.g. classification) (25, 30). The models were selected based on their average performance on the MTEB and the model size, opting for smaller model sizes that still provide strong results. To test the effect of embedding model diversity, we included variants of three of the selected models, selecting variants that are adapted to different tasks. Table S3 in the SI Appendix describes the models and their characteristics.

To map the categories from the coding scheme (22) onto the embedding space, the research team who performed the original coding selected 37 responses to represent the three primary codes (L, P, and S): 20 for *Limitations*, 7 for *Principles*, and 10 for *Statistics*. In addition, the team selected 22 responses to define the *Other* category, bringing the total number of category-defining responses to 59, representing approximately 2% of the dataset. These selected responses were chosen to capture the essence of their categories. For example, the code *Limitations* is defined as "Variability owing to our inability to perfectly model and measure a physical system in a real-world experiment"; representative responses included "Environmental influences" and "Ball not placed at the right spot each time." The embedding vectors of the selected responses for each category were averaged to create four category centroids.

Next, in the fourth step, we measure the distances between each of the 2899 embedded responses and each category centroid. In the present analysis, the coding scheme is mutually exclusive. Therefore, each response was assigned the category corresponding to its nearest category centroid.

To evaluate the classification performance of this framework, we benchmark these new classifications against those provided by the human coders, which we take as our "ground truth". In addition, we evaluate how the selection of responses used to define each category affects model performance (see the SI Appendix). We compute the F1 score ($F_1$), Cohen's Kappa ($\kappa$), and the multiclass Matthew's Correlation Coefficient (MCC) between the "ground truth" and the predicted codes.

**Dataset Audit Using Embeddings.** In the process of applying this framework to the coded data, we became aware of a number of coding inconsistencies. Some variation in coding consistency is a natural artifact of the process of manually qualitatively analyzing thousands of responses over a period of days or weeks. However, these inconsistencies made it difficult to benchmark our framework's performance, since in some cases identical responses had received two different codes. Therefore, we used embeddings to perform an audit of the dataset prior to testing our framework.

To do so, we measured the cosine distance between all pairs of responses and flagged responses that had at least one other response within a certain distance threshold (0.15) that had received a different code. This revealed that 669 responses were nearly identical in content or phrasing to some other response with a different code. For instance, the response "Distance between the slit and the screen" had been coded as L, while the response "The distance from the slit to the screen" had been coded as P. Such inconsistencies would be difficult to identify using simple keyword matching or other manual checks, but were easily identified using pairwise semantic distance measures.

The flagged responses were reviewed by the team that had performed the manual coding, and 531 of the inconsistencies were resolved by reclassifying 153 of the responses. All reported results are therefore on the audited dataset; results on the non-audited dataset can be found in the SI Appendix, see Tables S4-S8.

**Performance on a Simulated Exhaustive Coding Task.** As a first proof-of-concept, we test the framework's performance on a subset of the dataset that represents an example of mutually-exclusive, exhaustive coding, where every response gets a single code and no response is left out or coded as *Other*. To simulate this type of task, we drop the 410 responses that have been human coded as *Other* leaving 2,489 coded responses in the dataset.

The framework's performance on this scenario is presented in Table 2. Although there was some variation in performance across different embedding models, all showed very high levels of agreement with human coders across all three evaluation metrics. For example, Cohen's Kappa scores vary from 0.74 to 0.83. Kappa values above 0.61 are commonly recognized as substantial agreement, and values above 0.80 as almost perfect agreement (31). The standard deviation across the different resampling runs hovers around 0.03. Table S9 in the SI Appendix provides additional details on these results.

**Table 2. Results for deductive survey classification on a simulated exhaustive coding task by 6 different embedding models, plus variations. Exhaustive coding was simulated by removing responses coded as *Other* from the dataset, leaving only responses that had received a primary code.**

| Model | Instruction | $F_1$ | $\kappa$ | MCC |
|---|---|---|---|---|
| **Mixedbread large** | None | 0.95 | 0.77 | 0.77 |
| **Nomic v1** | None | 0.95 | 0.78 | 0.78 |
|  | Classification | 0.95 | 0.77 | 0.78 |
| **Jina small v2** | None | 0.95 | 0.74 | 0.74 |
| **Jina base v2** | None | 0.96 | 0.79 | 0.79 |
| **Jina v3** | None | 0.96 | 0.78 | 0.79 |
|  | Classification | 0.96 | 0.78 | 0.79 |
|  | STS | 0.96 | 0.79 | 0.79 |
| **Infloat large instruct** | None | 0.95 | 0.75 | 0.76 |
|  | STS | 0.96 | 0.83 | 0.83 |

Instruction refers to what was appended before all responses, None means no text was added. For Classification the added text was "classification: ". For STS (Semantic text similarity) the phrase "Instruct: Retrieve semantically similar text \n Query: " was appended (30).

**Performance on a Selective Coding Task Before and After Fine-tuning.** The second scenario we tested was mutually-exclusive coding with selective codes, where every response receives a single code but responses can be coded as *Other*. This is the approach used by the human coders in the manual analysis of the survey data, so we include the entire dataset. The results of this analysis are shown in Table 3. As shown, performance was significantly lower on this task than on the simulated exhaustive coding task, with Kappa values of 0.38-0.46, a range that is typically interpreted as fair to moderate agreement (31). The standard deviation when sampling is 0.02 for most models; see Table S10 in the SI Appendix for details.

What can explain this discrepancy in performance? In selective coding tasks, the *Other* category is usually broad and includes a combination of noise and groups of outliers — that is, responses that do not fit well into any of the primary categories. Such categories are, by definition, defined by what they are not. This presents a challenge for the proposed framework, as the fundamental principle of this method is to define categories by what they are (via examples). This makes it difficult to find a semantic vector or centroid for the *Other* category. Figure 3, which shows a 2D visualization of the embedded responses using t-SNE (32) colored by code, illustrates this difficulty; it can be seen that the responses labeled as *Other* (green diamonds) are scattered across the space and have a high overlap with the largest coded category, *Limitations*.

**Table 3. Results for selective coding before fine-tuning. Same experimental setup as before.**

| Model | Instruction | $F_1$ | $\kappa$ | MCC |
|---|---|---|---|---|
| **Mixedbread large** | None | 0.76 | 0.46 | 0.49 |
| **Nomic v1** | None | 0.74 | 0.43 | 0.47 |
|  | Classification | 0.74 | 0.43 | 0.47 |
| **Jina small v2** | None | 0.72 | 0.38 | 0.42 |
| **Jina base v2** | None | 0.74 | 0.41 | 0.45 |
| **Jina v3** | None | 0.76 | 0.45 | 0.48 |
|  | Classification | 0.75 | 0.44 | 0.47 |
|  | STS | 0.74 | 0.43 | 0.47 |
| **Infloat large instruct** | None | 0.76 | 0.46 | 0.49 |
|  | STS | 0.76 | 0.46 | 0.49 |

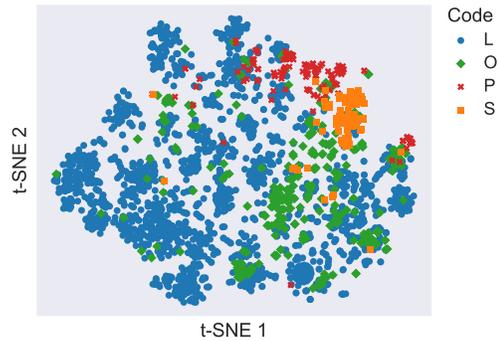

**Fig. 3.** 2-dimensional t-SNE plot of responses embedded with Jina small v2. Responses have been colored according to the category assigned by a human researcher.

Furthermore, the open-source embedding models used so far are generalists, trained to be adept at separating and comparing textual data across many different domains (33). As the *Other* category by definition lacks a central semantic theme, it seems unlikely that a standard pre-trained embedding model can be expected to group arbitrary responses neatly without additional training.

To begin to address these theoretical limitations, we have taken the additional step of re-analyzing the complete dataset, but this time fine-tuning the embedding models. Fine-tuning is a process in which an embedding space can be reshaped to fit the needs of a specific task. When fine-tuning, researchers provide labeled examples of texts, some of which are specified to be close and others far apart in the semantic space (training data) (34). Then, using a learning function, the geometry of the embedding space is updated to better fit these examples. In principle, this functionality makes it possible to manipulate the embedding space to create an *Other* "area".

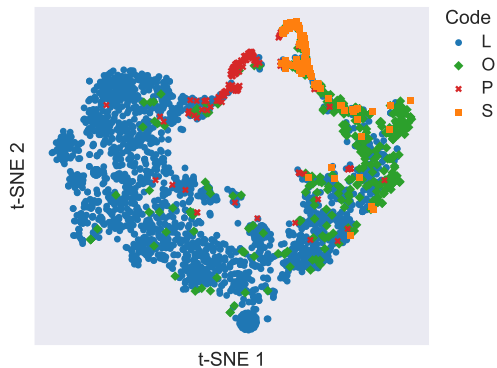

**Fig. 4.** 2-dimensional t-SNE plot of responses embedded with Jina small v2 after fine-tuning. Responses have been colored according to the category assigned by a human researcher (after data audit).

To apply this principle here, we create training data by labeling the responses that were used to define the category centroids. To create a slightly larger dataset without changing the models ratio of seen to unseen data, we augmented the 59 category-defining responses with 34 additional responses drawn from a related but independent dataset consisting of responses from the same survey, but to a different question: see the SI Appendix for details. These 34 responses were categorized into the same four categories (21). This resulted in a fine-tuning training dataset of 93 texts (28 *Limitations*, 15 *Principles*, 19 *Statistics*, 31 *Other*). From these 93 responses, we generated all ordered pairs (93 × 93 = 8,649), labeling each pair with 1 if both responses shared the same category and 0 otherwise.

We then fine-tuned all the six standard text embedding models and again classified the entire dataset. Note, that in this analysis no embedding models were given instructions; additional details on the fine-tuning procedures used can be found in the SI Appendix. Results are summarized in Table 4. The results show a substantial improvement between ten and twenty percent on both Kappa and MCC scores for all models. Examination of the t-SNE plot of responses in a fine-tuned embedding space (shown in Figure 4) provides some explanation for this improvement. As can be seen, after fine-tuning the regions corresponding to different codes have become significantly more distinct. In particular, responses coded as *Other* have been grouped more strongly, although there is still some overlap between those responses and the responses coded as *Limitations*.

**Table 4.** Results for selective coding after fine-tuning. Same experimental setup as before.

| Model | Instruction | $F_1$ | $\kappa$ | MCC |
|---|---|---|---|---|
| **Mixedbread large** | None | 0.80 | 0.53 | 0.57 |
| **Nomic v1** | None | 0.78 | 0.50 | 0.52 |
| **Jina small v2** | None | 0.75 | 0.50 | 0.52 |
| **Jina base v2** | None | 0.80 | 0.53 | 0.55 |
| **Jina v3** | None | 0.80 | 0.52 | 0.54 |
| **Infloat large instruct** | None | 0.79 | 0.51 | 0.53 |

## Discussion

In this paper, we have proposed a framework for using embeddings to qualitatively analyze open-ended survey data at scale. This framework fits well into standard qualitative workflows, since it overlaps significantly with the way deductive coding is typically done by human researchers and requires only a small number of example responses to run.

When applied to a dataset consisting of thousands of free-text responses, the proposed framework shows consistently strong performance on a simulated exhaustive coding task (where every response received one of the three primary codes). Across ten different embedding models, classifications had strong agreement with the codes assigned by a human researcher, with Cohen's Kappa scores in the range of 0.74 to 0.83. These results indicate that for this type of task, the proposed framework reliably replicates expert coding decisions based on a limited number of example responses. Furthermore, by identifying neighboring responses in the semantic space which had received different human-assigned codes, we were able to audit the dataset for coding inconsistencies and edge-cases.

However, performance was lower when applying this framework to the full, selectively-coded dataset that excluded certain responses (by grouping them into the category of *Other*). These limitations were somewhat mitigated by fine-tuning the embeddings models, but performance still remains far below that in the simulated exhaustive coding task. Because the *Other* category by definition lacks a central semantic theme, we argue that the proposed framework will likely continue to struggle on these types of coding tasks.

Both the study and the framework have several limitations. In the present study, one key limitation comes from the fact that the analysis was performed on a single dataset, from a specific domain (physics education). It is unclear how the topic or quality of this dataset has influenced the results. For instance, responses in the dataset were typically quite short, meaning that they likely focused on a single theme or idea; longer or more elaborate responses may perform differently or require alternative analysis approaches, such as segmenting responses on a sentence-by-sentence level and then individually categorizing those sentences.

Second, the categories used in the analysis were fairly distinct. There is reason to believe that the framework will perform more poorly a more diverse dataset, with theoretical categories that require inference on the part of a human researcher. This is because text embedding models represent semantic meaning of text; they are therefore well-suited for content analysis, but likely less suited for analysis of psychological or motivational constructs that are identified based on latent or gestalt textual features using human inference.

Finally, there is reason to believe that the fine-tuning procedure used in this study would improve with careful hyper-parameter tuning, as the standard library parameters that we used are likely suboptimal for most models. However, to avoid scope-creep in the present study we have limited our exploration of fine-tuning to a simple proof-of-concept.

Theoretically, there are also challenges when applying this framework to analysis of datasets that have not been previously analyzed or coded. For starters, when using the framework on a new dataset, it will be significantly harder to evaluate or benchmark the results, absent some amount of additional analysis by a human coder. It is also unclear how to choose a specific embedding model for such a task, or to systematically predict the effects of varying the chosen example texts on the results.

Despite these challenges, this framework remains a promising avenue of research for expanding the scope of survey-based qualitative studies. Specifically, the proposed embedding-based approach has the benefit of simplicity, transparency, and replicability. The simplicity comes from the fact that the method (both classification and fine-tuning) can be run using a limited number of example responses, on arbitrarily large datasets. This type of analysis is also more transparent and replicable than standard qualitative analysis, in that the machine will always code identical responses the same way, and data, examples, and analysis code can easily be shared and run by other researchers to evaluate the quality of results. Simultaneously, the proposed method for auditing datasets can be used on both machine-coded and human-coded data, which represents a promising addition to the existing toolkit for identifying errors and inconsistencies in qualitative datasets (35).

We emphasize that one of the key benefits of the proposed framework is that the qualitative researcher remains an integral part of the process. Whereas other NLP methods often struggle to strike a balance between efficiency and transparency, our approach places qualitative researchers right where they belong: at the heart of the analysis. Furthermore, because researchers are able to flexibly redefine categories by picking new category-defining texts, the proposed framework has the potential to significantly speed up data exploration.

Additionally, we note that several of the challenges and limitations identified above are equally present, if not more so, in traditional "by hand" qualitative survey analysis. Humans also struggle to analyze diverse datasets that require significant amounts of inference, and the difficulties with benchmarking analyses on new datasets are equally present for human coders. Furthermore, when coding schemes are applied selectively it can be difficult to determine criteria for inclusion or exclusion into a specific category, requiring judgments that may vary when qualitative coding tasks extend over hours, days, or weeks.

Based on these challenges and affordances, we identify several paths of potential future research. First, we see the need to test this framework on other datasets, especially those that have been previously coded by a human researcher, to empirically determine areas of applicability. Second, we see a need to adapt this framework to different kinds of qualitative coding schemes, especially multiclass schemes that are not mutually exclusive. Mapping these types of coding tasks onto the realm of embeddings represents a significant theoretical challenge. Finally, the technology underlying embeddings is rapidly developing, and new models are released nearly every day. Many of these models have unexplored capabilities that may further enhance qualitative analysis, such as natively embedding image data alongside text. However, in addition to exploring their applications to qualitative analysis, more research needs to be done on how these models work, and why they work as they do. Increased understanding of embeddings, combined increased familiarity with their applications, may lead to better utilization of the vector spaces and more powerful applications within qualitative survey analysis.

**Materials and Methods**

Summary data are provided in the article and SI Appendix. The main dataset, including raw responses and codes, is available at Zenodo (https://doi.org/10.5281/zenodo.16912394). All analyses were conducted in Python using code available at GitHub (https://github.com/halvorty/text2embedding2category).

**ACKNOWLEDGMENTS.** This work was funded by the Norwegian Directorate for Higher Education and Skills (HK-dir), which supports the University of Oslo's Center for Computing in Science Education and Center for Interdisciplinary Education. This material is based upon work supported by the National Science Foundation Grants No. DUE-2336135 and DUE-2336136.